\documentclass[letterpaper, 10 pt, conference]{ieeeconf}

\IEEEoverridecommandlockouts                              

\usepackage{graphicx} 
\usepackage{epsfig} 
\usepackage{dblfloatfix} 
\usepackage{courier}
\usepackage{amsmath} 
\usepackage{amssymb}  
\usepackage{amsmath} 
\usepackage{amssymb}

\usepackage{float}
\usepackage{subfig}
\newlength{\subfigheight}
\newsavebox{\subfigbox}

\usepackage{multirow}
\usepackage{lipsum}
\usepackage{siunitx}

\makeatletter
\AtBeginDocument{
  \check@mathfonts
}
\makeatother

\usepackage[backend=bibtex,bibstyle=ieee,citestyle=numeric-comp,isbn=false,doi=false]{biblatex}
\title{\LARGE \bf
Performance Evaluation of 3D Keypoint Detectors and Descriptors on Coloured Point Clouds in Subsea Environments
}

\author{Kyungmin Jung, Thomas Hitchcox, James Richard Forbes\\  
\thanks{This work was supported in part by Voyis Imaging Inc. through the NSERC CRD Program and in part by the MEDA Program at McGill University. The authors are with the department of Mechanical Engineering, McGill University, Montreal, QC H3A 0C3, Canada. {\{\tt\small kyungmin.jung@mail.mcgill.ca, thomas.hitchcox@mail.mcgill.ca, james.richard.forbes@mcgill.ca\}.}}}

\newcommand{\ignore}[1]{}

\newcommand{\norm}[1]{\left\Vert#1\right\Vert} 
\newcommand{\abs}[1]{\left\vert#1\right\vert} 
\newcommand{\mc}[1]{\mathcal{#1}}

\newcommand{\bma}[1]{\left[\begin{array}{ #1}}
\newcommand{\ema}{\end{array}\right]}

\DeclareMathAlphabet{\mbf}{OT1}{ptm}{b}{n}
\newcommand{\mbs}[1]{{\boldsymbol{#1}}}

\newcommand{\mbfhat}[1]{{\hat{\mbf{#1}}}}
\newcommand{\mbfcheck}[1]{\ensuremath{\check{\mbf{#1}}}}

\newcommand{\mbftilde}[1]{{\tilde{\mbf{#1}}}}

\def\fdotb{{\raisebox{-0.6ex}{ \kern0.2ex\raisebox{0.8ex}{\tiny $\hspace*{-1ex}\circ$}}}}
\def\fddotb{{\raisebox{-0.6ex}{ \kern0.2ex\raisebox{0.8ex}{\tiny $\hspace*{-1ex}\circ\circ$}}}}

\newcommand{\f}{\frac}

\newcommand{\ura}[1]{{\underrightarrow{{#1}}}}

\newcommand{\trans}{{\ensuremath{\mathsf{T}}}} 
\newcommand{\utimes}{ {\raisebox{-0.6ex}{ \kern-1.0ex\raisebox{0.6ex}{ \small $\mathsf{v}$}}} } %
\newcommand{\beq}{\begin{equation}}
\newcommand{\eeq}{\end{equation}}
\newcommand{\bdis}{\begin{displaymath}}
\newcommand{\edis}{\end{displaymath}}
\newcommand{\beqarray}{\begin{eqnarray}}
\newcommand{\eeqarray}{\end{eqnarray}}
\newcommand{\beqarraynn}{\begin{eqnarray*}}
\newcommand{\eeqarraynn}{\end{eqnarray*}}
\newcommand{\balign}{\begin{align}}
\newcommand{\ealign}{\end{align}}
\newcommand{\balignnn}{\begin{align*}}
\newcommand{\ealignnn}{\end{align}}

\renewcommand{\theenumii}{\arabic{enumii}}
\makeatletter
\renewcommand{\p@enumii}{\theenumi.}
\makeatother

\usepackage{arydshln}

\begin{document}
\fontdimen16\textfont2=\fontdimen17\textfont2
\fontdimen13\textfont2=5pt

\maketitle
\thispagestyle{empty}
\pagestyle{empty}

\begin{abstract}
    The recent development of high-precision subsea optical scanners allows for 3D keypoint detectors and feature descriptors to be leveraged on point cloud scans from subsea environments.
    However, the literature lacks a comprehensive survey to identify the best combination of detectors and descriptors to be used in these challenging and novel environments.
    This paper aims to identify the best detector/descriptor pair using a challenging field dataset collected using a commercial underwater laser scanner.
    Furthermore, studies have shown that incorporating texture information to extend geometric features adds robustness to feature matching on synthetic datasets.
    This paper also proposes a novel method of fusing images with underwater laser scans to produce coloured point clouds, which are used to study the effectiveness of 6D point cloud descriptors.
\end{abstract}
\section{Introduction}
Conventional vision-based navigation has seen ample application in the field of subsea robotics.
However, underwater environments pose unique challenges to vision-based navigation algorithms, such as light scattering, poor and inconsistent illumination, and feature-depleted scenes \cite{Rahman2019SVIn2}.

Newly developed high-precision subsea 3D optical scanners allow for more robust feature extraction in subsea environments \cite{Castillon2019State}.
For example, feature matching techniques have successfully been leveraged to initialize the alignment of bathymetric point cloud scans via a coarse alignment process \cite{Palomer2019Inspection}.
This approach has been effective in bounding the navigation drift over time when used within an underwater simultaneous localization and mapping (SLAM) framework.

However, an evaluation of 3D keypoint detectors and descriptors for subsea environments appears to be absent in the literature.
Moreover, a study on which detector/descriptor pairs work best in terms of robustness and reliability when used to initialize an iterative closet point (ICP) pipeline in subsea environments has yet to be thoroughly explored.
This paper surveys the performance of 3D detector/descriptor pairs for point cloud alignment in a variety of underwater environments.
The motivation for such a study stems from the need to incorporate loop closures within a bathymetric SLAM pipeline to correct odometry drift.

Furthermore, studies on synthetic datasets have shown that incorporating colour information to extend geometric descriptors improves the robustness of feature matching \cite{Tombari2011CSHOT}.
As a secondary contribution, this paper proposes a novel method of colouring point clouds using an underwater laser scanner and an external colour camera.
The resulting coloured point clouds are used to evaluate the performance of colour-enhanced feature descriptors.

In detail, this paper's contributions include
\begin{itemize}
    \item generation of 3D coloured point clouds using an underwater laser line scanner and an external colour camera to add robustness to feature extraction,
    \item a robustness analysis of various 3D keypoints subject to different transformations and point cloud noise levels in different underwater environments, and
    \item a performance evaluation of 3D features in initializing an ICP fine alignment algorithm.
\end{itemize}

The remainder of this paper is organized as follows.
Section~\ref{sec:RelatedWork} begins by outlining related work.
Section~\ref{sec:CPCG} presents the novel method of colour point cloud generation.
Section~\ref{sec:Methodology} then explains the methods used to evaluate various detector/descriptor pairs with coloured underwater point clouds.
Section~\ref{sec:Results} presents the evaluation results.
Finally, Section~\ref{sec:Conclusion} summarizes the findings.
\begin{figure}[tb]
    \centering
    \subfloat[Feature Matching\label{cover_image_a}]{%
        \includegraphics[width=0.5\linewidth]{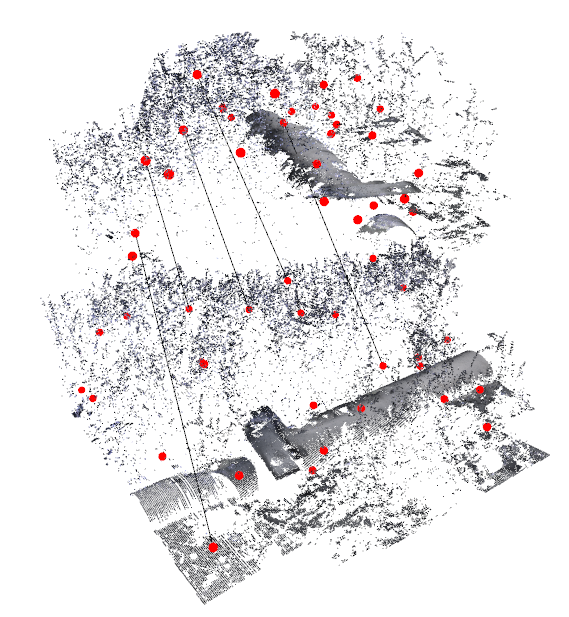}}
    \subfloat[Aligned Submap\label{cover_image_b}]{%
        \includegraphics[width=0.5\linewidth]{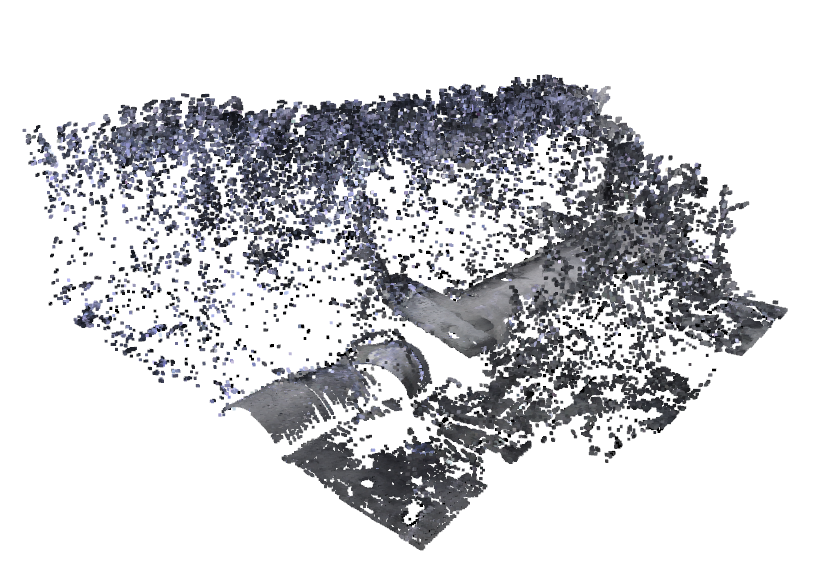}}
    \caption{Point cloud alignment via feature matching is shown here.
        Fig~\ref{cover_image_a} shows two point cloud submaps generated from laser scans of a subsea pipeline.
        Note the submaps are separated in the vertical direction for easier visualization.
        The extracted keypoints from the two submaps are shown as red dots.
        The aligned submaps are shown in Fig~\ref{cover_image_b}.}
    \label{fig:cover_image}
\end{figure}
%

\section{Related Work}
\label{sec:RelatedWork}
The effectiveness of different detector/descriptor pairs has been evaluated in the literature via terrestrial benchmark datasets.
For example, a comprehensive evaluation of state-of-the-art 3D keypoints was conducted in \cite{Salti2011Performance}, which mainly addressed their common traits in 3D object recognition tasks.
The evaluation was carried out to analyze repeatability and computational efficiency using synthetic datasets from \cite{Curless1996Stanford}.
Keypoint repeatability was measured as a detector's ability to find the same set of keypoints on different instances of a given point cloud that may be corrupted by noise.
Overall, KPQ-SI, MeshDoG, and ISS yielded the best scores in terms of repeatability in simulated data.
Further, \cite{Salti2011Performance} stated that the accuracy of feature matching could be improved by incorporating texture information into the descriptors.

The descriptiveness and robustness of ten 3D local descriptors were studied in \cite{Guo2016Comprehensive}.
The descriptiveness was evaluated using a precision-recall curve, and the robustness was evaluated with respect to varying support radius, noise values, and mesh resolutions.
However, \cite{Guo2016Comprehensive} analyzed the performance in terms of object recognition and shape retrieval and did not consider navigation applications.

More recently, \cite{Han2018Point} presented a comprehensive study of 3D point cloud descriptors, including more recently developed deep-learning-based descriptors.
The paper also described the traits of local descriptors, which construct a histogram around keypoints within the input point cloud, and global descriptors, which generally estimate a single descriptor vector encoding the whole input cloud for scenarios like 3D object recognition and geometric categorization.

\section{Coloured Point Cloud Generation}
\label{sec:CPCG}
This section discusses data collection methods, point cloud submap generation, and the novel method of point cloud colourization using a high-precision underwater laser line scanner and an external RGB camera.

\subsection{Underwater Field Dataset}
\label{subsec:Dataset}
\begin{figure}[tb]
    \sbox\subfigbox{%
        \resizebox{\dimexpr\columnwidth}{!}{%
            \includegraphics[height=4cm]{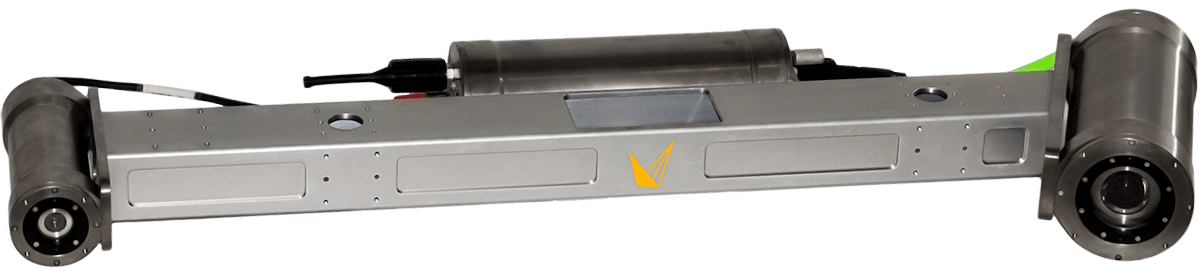}%
        }%
    }
    \setlength{\subfigheight}{\ht\subfigbox}
    \centering
    \includegraphics[height=0.75\subfigheight]{figs/voyis-uls-500-pro.png}
    \par \smallskip
    \includegraphics[height=0.75\subfigheight]{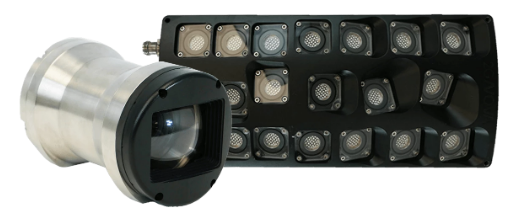}
    \caption{A laser line scanner, \textit{Insight Pro} (top), and an imaging system, \textit{Observer Micro} (bottom), provided by Voyis Imaging Inc. are used to collect point cloud and image data, respectively.}
    \label{fig:hardware}
\end{figure}
%
Field data was collected at Colpoy's Bay, Ontario, Canada by Voyis Imaging Inc. using their surface vessel equipped with a Sonardyne SPRINT-Nav 500 Doppler velocity log-aided inertial navigation system (DVL-INS), a u-blox ZED-F9P high precision GNSS module equipped with a u-blox ANN-NM series high precision multi-band antenna, a Voyis \textit{Insight Pro} underwater laser scanner, and an \textit{Observer Micro} underwater imaging system.
The laser scanner and the imaging system are shown in Fig~\ref{fig:hardware}.
Laser `submaps' are constructed by registering individual laser line profiles to the trajectory estimate produced by the DVL-INS.
Although the DVL-INS provide very accurate navigation estimate, its drift becomes significant in long-term missions.
A detailed explanation of point cloud generation and colourization is explained in Section~\ref{subsec:SubmapRegistration} and Section~\ref{subsec:PointCloudColourization}, respectively.

\begin{figure}[tb]
    \centering
    \includegraphics[width=\linewidth,clip=true,trim={0.75cm 2.75cm 1.5cm 3.75cm}]{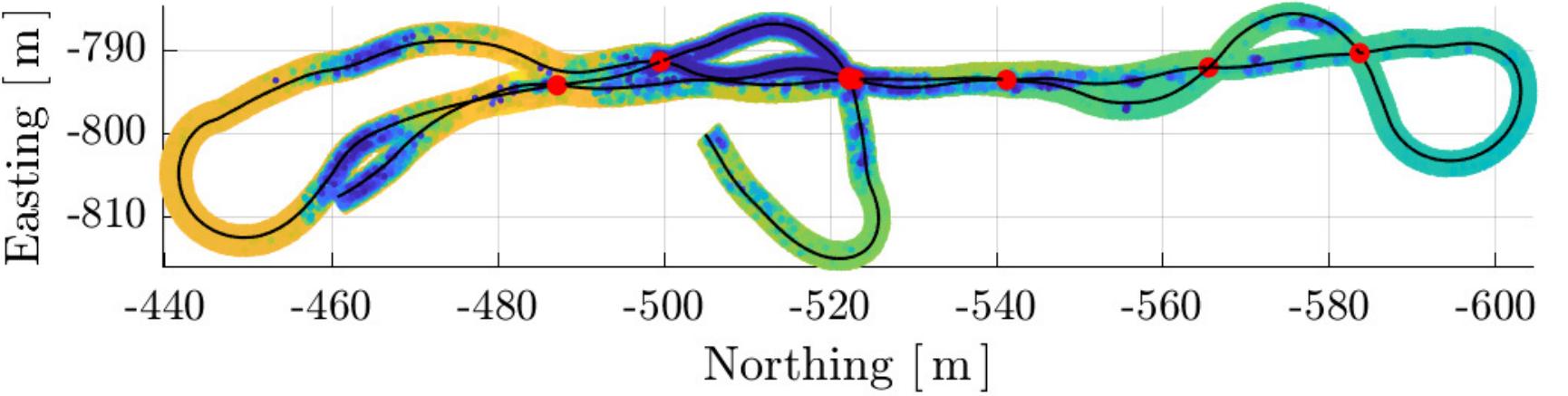}
    \caption{The vehicle trajectory over a pipeline target is shown as a black line, while loop-closure crossings are shown as red circles.
        The colour of the underlying point cloud scan represents relative depth across the dataset.}
    \label{fig:traj_lc}
\end{figure}

The collected field data is divided into four sections: two shipwreck sections, a pipeline section, and a seabed section.
Each section of the data consists of multiple loop-closure events as the vehicle makes a loop around a previously visited region as shown in Fig~\ref{fig:traj_lc}.
For each loop closure, a pair of submaps is generated around the two intersecting vehicle poses.
Each submap pair is qualitatively assigned to one of three categories depending on the local environment.
These categories are structured scenes such as shipwrecks and pipelines, semi-structured scenes such as debris fields, and unstructured scenes such as seabed and marine plant life.
The number of classified submaps is shown in Table~\ref{tab:dataset}.
\begin{table}[tb]
    \centering
    \caption{A summary of the field data in number of loop closures.}
    \label{tab:dataset}
    \begin{tabular}{c | c c c c}
        Section     & Structured & Semistructured & Unstructured & Total \\
        [1ex]
        \hline
        Shipwreck A & 9          & 6              & 6            & 21    \\
        Shipwreck B & 13         & 16             & 54           & 83    \\
        Pipeline    & 6          & 2              & 0            & 8     \\
        Seabed      & 0          & 0              & 6            & 6     \\
        Total       & 28         & 24             & 66           & 118
    \end{tabular}
\end{table}

\subsection{Submap Generation}
\label{subsec:SubmapRegistration}
The Voyis \textit{Insight Pro} laser line scanner measures a single array of points ${\mc{P}_{\ell} = \{\mbf{r}^{p_{i}s}_{\ell}\}_{i=1}^{N}}$, where $\mbf{r}^{p_{i}s}_{\ell}$ is the position vector of reflected point $p_{i}$ relative to the sensor $s$ resolved in the laser frame $\mc{F}_{\ell}$, and $N$ is the total number of points in a single scan line.

A submap at time $t_{k}$ is composed of multiple nearby laser line measurements resolved in the vehicle body frame at time $\mc{F}_{b_{k}}$.
Thus, laser points measured at some time $t_{k}$ near a loop-closure time $t_{\tau}$ are transformed using the vehicle navigation estimate,
\begin{align}
    \label{eq:submap1}
    \mbf{r}^{p_{i}z_{\tau}}_{b_{\tau}} = \left(\mbf{T}^{z_{\tau}w}_{ab_{\tau}}\right)^{-1} {\mbf{T}^{z_{k}w}_{ab_{k}}} \mbf{T}^{sz}_{b\ell} \mbftilde{r}^{p_{i}s}_{\ell},
\end{align}
where $\mbftilde{r}^{p_{i}s}_{\ell}$ is the homogeneous form of $\mbf{r}^{p_{i}s}_{\ell}$, and
\begin{align}
    \mbf{T}^{z_{k}w}_{ab_{k}} = \begin{bmatrix}
                                    \mbf{C}_{ab_{k}} & \mbf{r}^{z_{k}w}_{a} \\ \mbf{0} & 1
                                \end{bmatrix} \in SE(3)
\end{align}
represents the pose of the vehicle as an element of matrix Lie group $SE(3)$.
Here, ${\mbf{r}^{z_{k}w}_{a} \in \mathbb{R}^{3}}$ denotes the position of the vehicle body, $z_{k}$, relative to some reference point $w$ resolved in the world frame $\mc{F}_{a}$ at some time $t_{k}$, and the direction cosine matrix ${\mbf{C}_{ab_{k}} \in SO(3)}$ represents the attitude of the body frame ${\mc{F}_{b_{k}}}$ at time $t_{k}$ relative to the reference frame $\mc{F}_{a}$.
Further, $\mbf{T}^{sz}_{b\ell}$ denotes the static extrinsics transformation from the laser sensor to the vehicle.
Denote ${\mc{P}^{k}_{b_{\tau}} = \{\mbf{r}^{p_{k,i}z_{\tau}}_{b_{\tau}}\}_{i=1}^{N_{k}}}$ as the laser line measurement at time $t_{k}$ resolved in the body frame at loop-closure time $t_{\tau}$.
Then a submap is constructed as
\begin{align}
    \mc{S}_{b_{\tau}} = \{ \mc{P}^{k}_{b_{\tau}} \}_{k=1}^{M},
\end{align}
where $M$ is the number of laser line measurements in the submap.
Following \cite{Roman2007SelfConsistent}, the size of the submap is chosen as $\SI{5}{\meter}\times\SI{5}{\meter}$ such that it is small enough to be assumed error-free and large enough to contain sufficient 3D information to be aligned to another submap.
Further, all submaps are downsampled using the voxel grid filter from \cite{Rusu2011PCL} with the grid size $\SI{0.05}{\meter}$.
The grid size is chosen to perform well with the dataset from Section~\ref{subsec:Dataset}.

An accurate laser-to-body extrinsics estimate is necessary for accurate point cloud registration \eqref{eq:submap1}.
The laser-to-body extrinsics are estimated a priori by minimizing the sum of squared reprojection errors.

\subsection{Coloured Point Clouds}
\label{subsec:PointCloudColourization}
\begin{figure}[tb]
    \centering
    \subfloat{%
        \includegraphics[width=0.495\linewidth,clip=true,trim={0cm 0cm 0.5cm 0.5cm}]{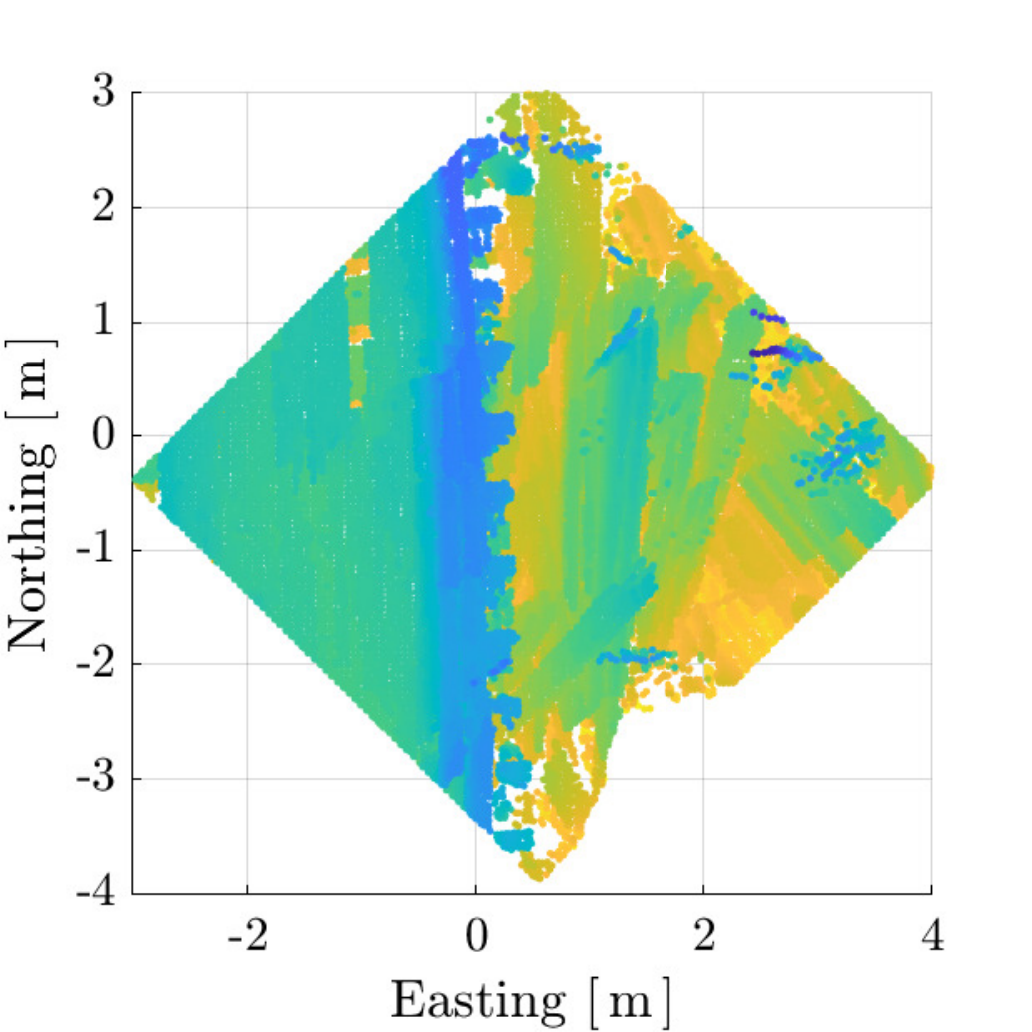}}
    \subfloat{%
        \includegraphics[width=0.495\linewidth,clip=true,trim={0cm 0cm 0.5cm 0.5cm}]{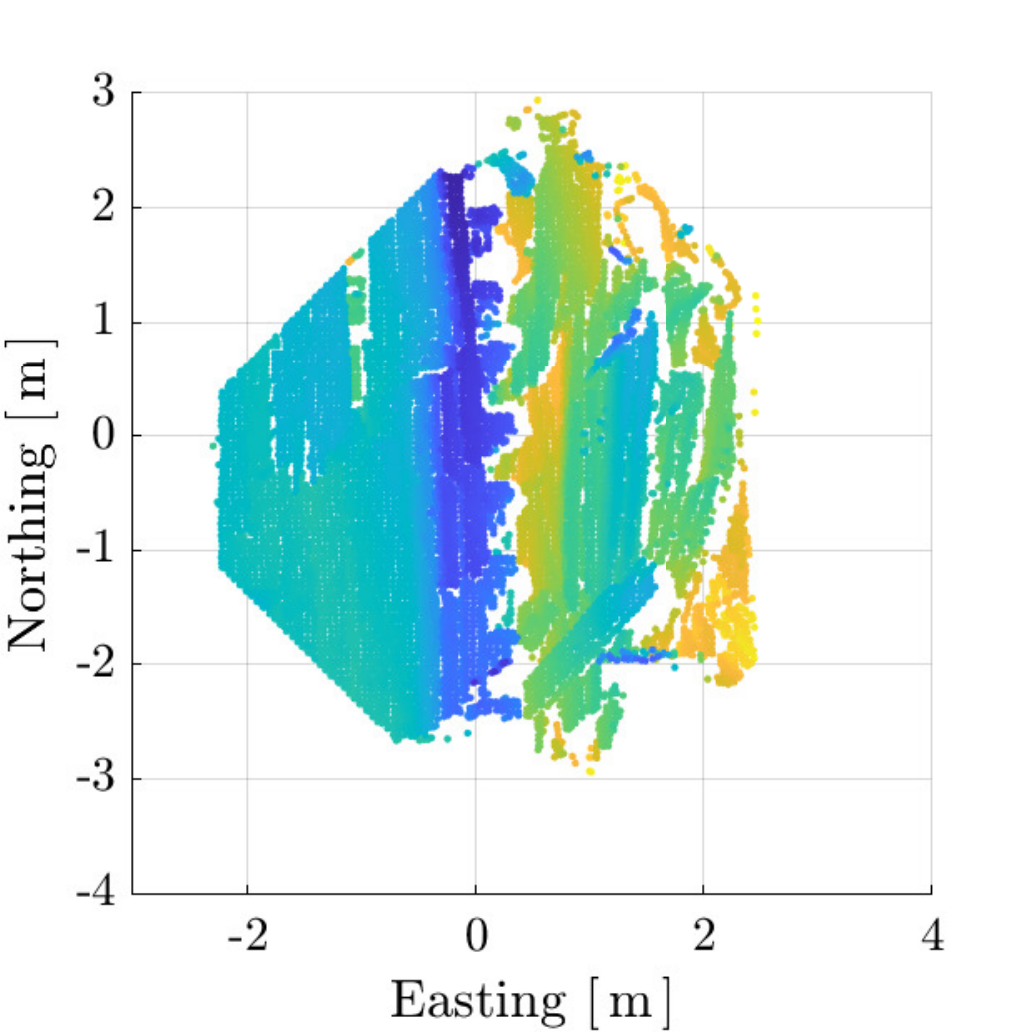}}
    \caption{
        A shipwreck is reconstructed from laser scans (left).
        Occluded points from the camera's viewpoint are removed using the method from \cite{Katz2007Direct} (right), so as to only add colour to points visible to the camera.
    }
    \label{fig:visibility}
\end{figure}
%
Incorporating texture information to extend geometric features has been shown to improve the feature matching accuracy \cite{Tombari2011CSHOT}.
The point clouds generated in Section~\ref{subsec:SubmapRegistration} are textured using images from an underwater colour camera.

To texture the submap, the point cloud is first transformed to a camera frame to determine point visibility. In particular, each $\mbf{r}^{p_{i}z}_{b}$ is transformed via
\begin{align}
    \mbftilde{r}^{p_{i}c}_{m} = \left(\mbf{T}^{cz}_{bm}\right)^{-1} \mbftilde{r}^{p_{i}z}_{b},
\end{align}
where ${\mbf{T}^{cz}_{bm} \in SE(3)}$ is the camera-body extrinsic matrix estimated using stereo calibration.
To determine the set of points visible to the camera, the approach presented in \cite{Katz2007Direct} is used, where a convex hull is constructed by transforming the point cloud to a new domain.
The points that lie on the convex hull of the transformed set of points are considered visible to the camera.
An example is shown in Fig~\ref{fig:visibility}.

After determining point visibility, the visible points are projected to the image plane using a pinhole camera model
\begin{align}
    \begin{bmatrix}
        u & v & w
    \end{bmatrix}^{\trans} & = \mbf{K} \, \mbf{r}^{p_{i}c}_{m},
\end{align}
where $\mbf{K}$ is the camera intrinsic matrix.
The colours from the corresponding pixels are assigned to the points.

\begin{figure}[tb]
    \centering
    \includegraphics[width=0.95\linewidth,clip=true,trim={0.3cm 0.3cm 0.3cm 0.3cm}]{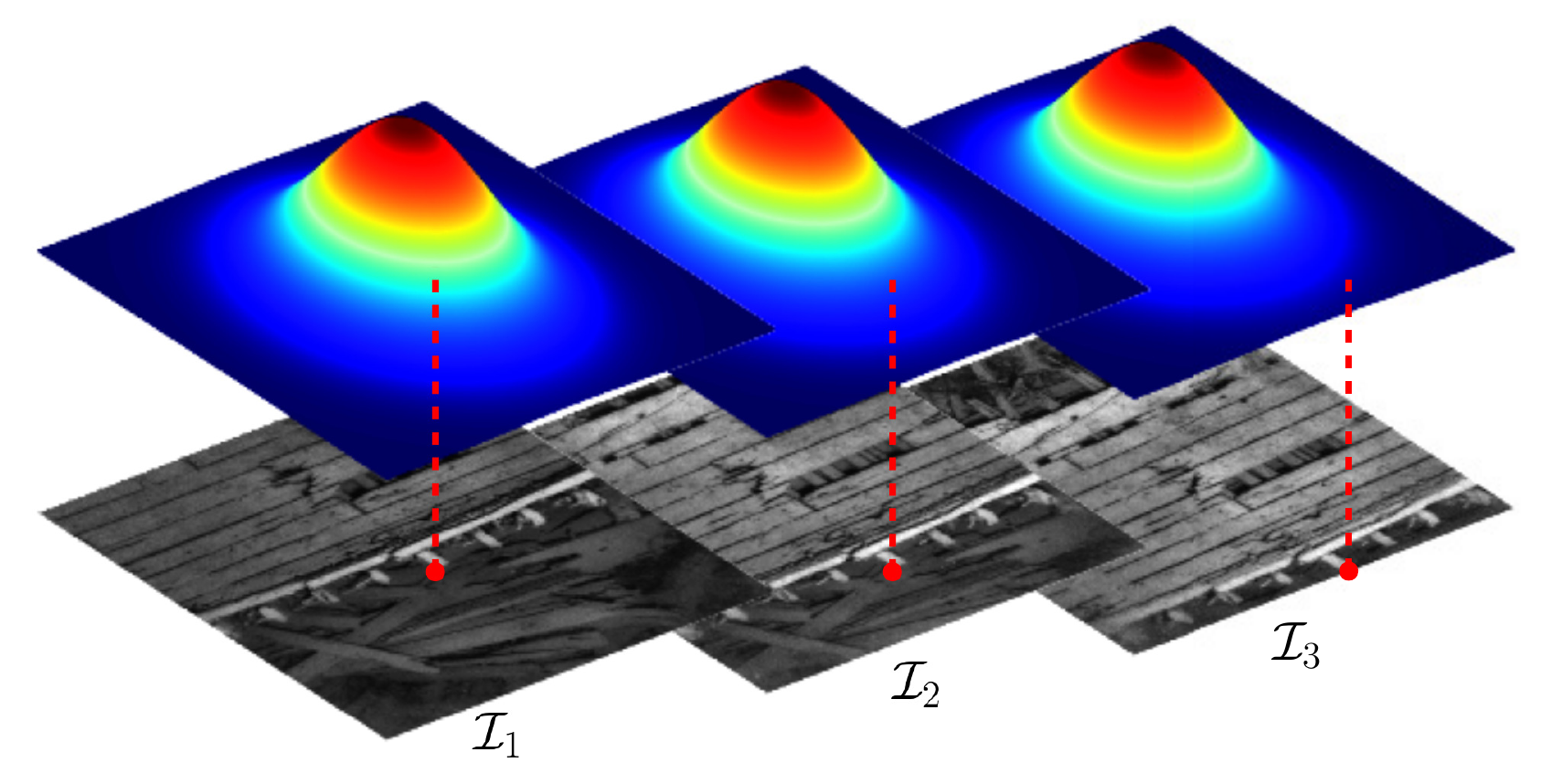}
    \caption{
        A 3D point projected to an image frame ${\mc{I}}$, shown as a red dot, is visible in multiple images at different pixel locations.
        Each colour candidate is weighted with the corresponding value of the 2D Gaussian.
        A unique colour is assigned by taking the sum of all weighted colour candidates \eqref{eq:colour}.
    }
    \label{fig:gaussian_weight}
\end{figure}

\begin{figure}[tb]
    \centering
    \subfloat{%
        \includegraphics[width=0.95\linewidth,clip=true,trim={0.5cm 0.25cm 0.5cm 1cm}]{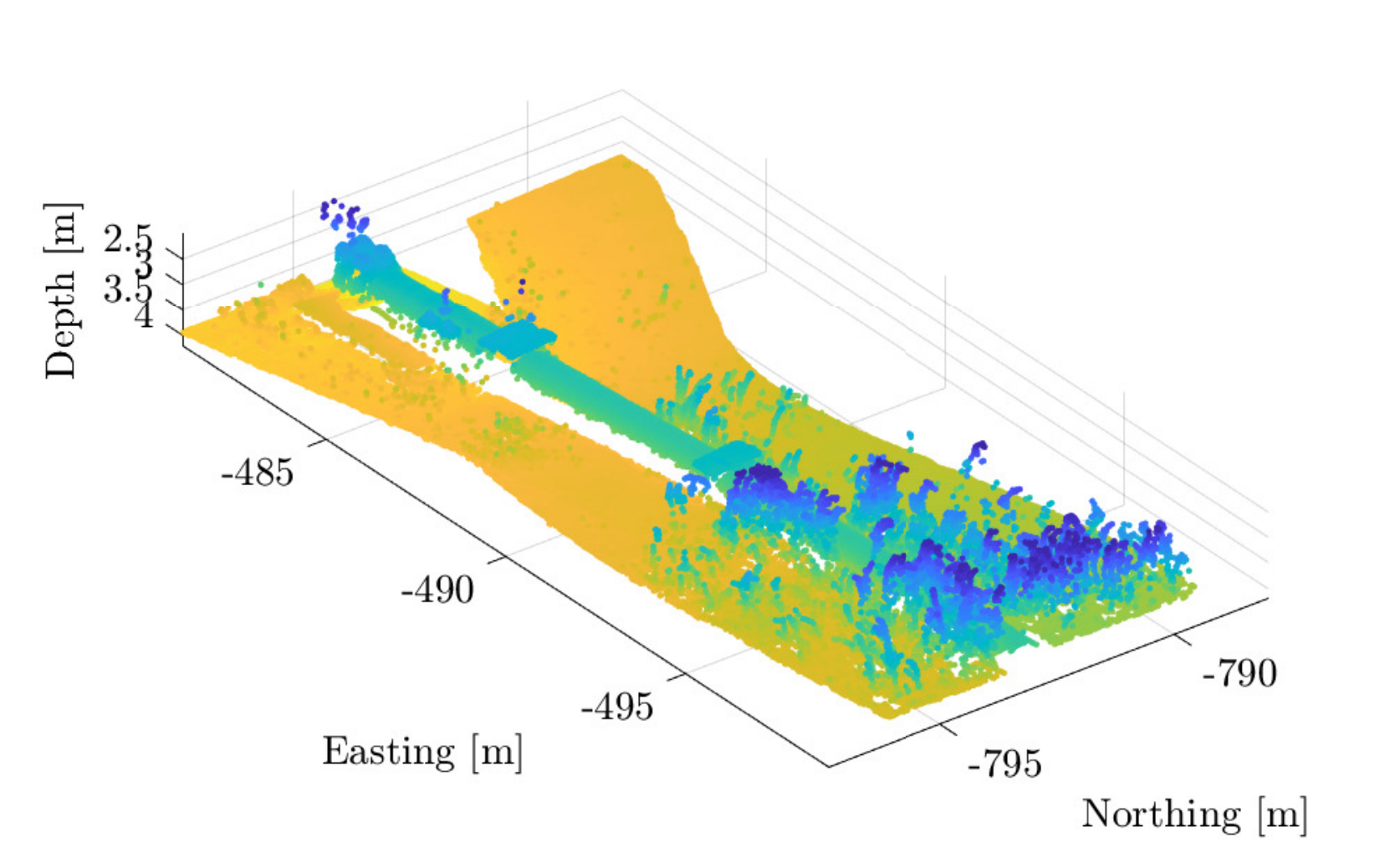}} \\
    \subfloat{%
        \includegraphics[width=0.95\linewidth,clip=true,trim={0.5cm 0.25cm 0.5cm 1cm}]{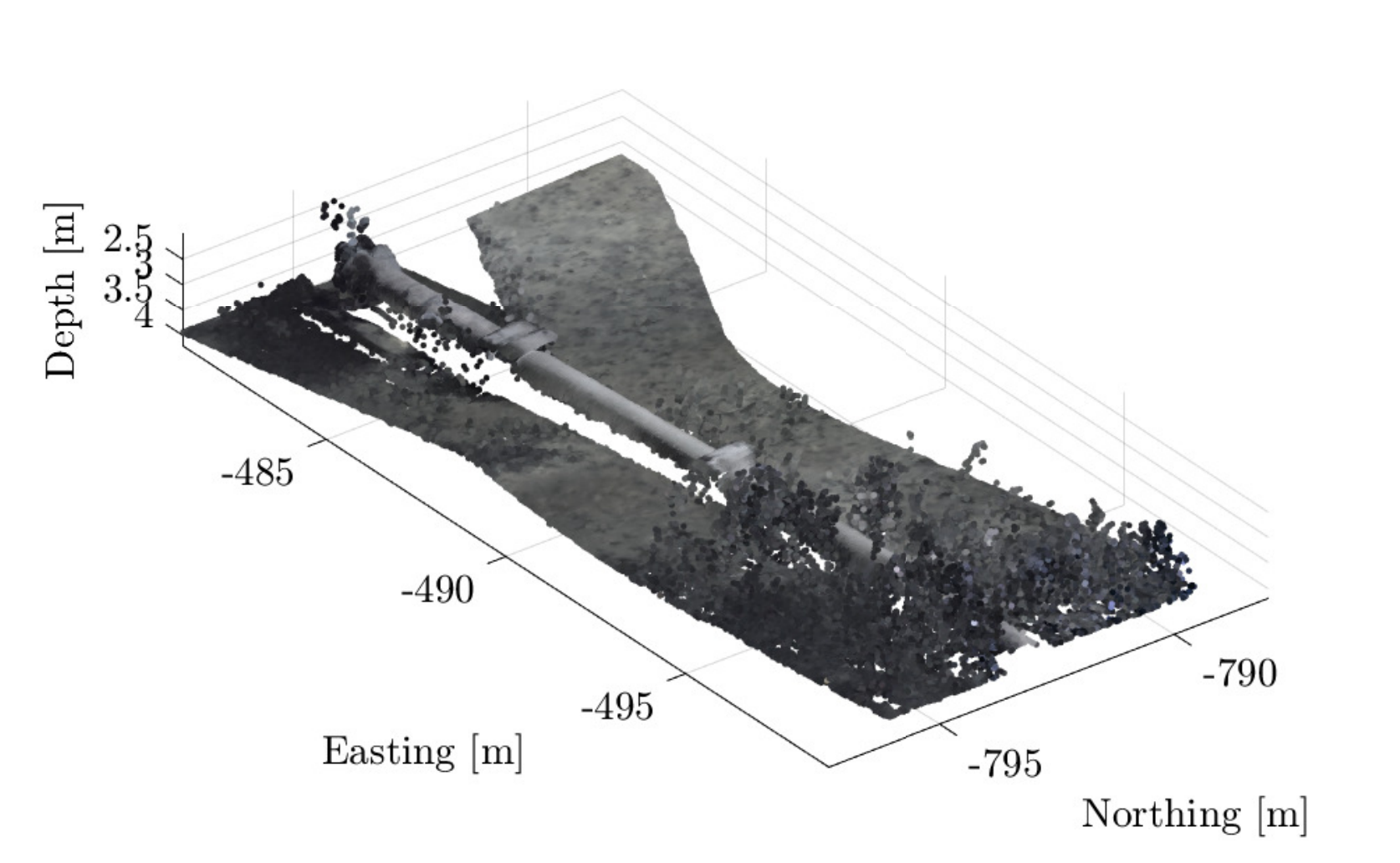}}
    \caption{A submap generated from the pipeline data displayed in elevation (top) and in colour (bottom).}
    \label{fig:coloured_map}
\end{figure}

In practice, the points in the submap may be seen in multiple camera frames at different observation angles resulting in multiple colour assignments.
This issue is solved in \cite{Vechersky2018Colourising} using a weighting scheme,
\begin{align}
    \label{eq:colour}
    \bar{x} = \f{\sum_{i}^{N_{c}} w_{i}x_{i}}{\sum_{i}^{N_{c}} w_{i}},
\end{align}
where $N_{c}$ is the total number of colour candidates, $x_{i}$ is the $i$-th colour candidate, and $w_{i}$ is the weight associated to each colour candidate.
The weights are chosen from a 2D Gaussian function overlaid on top of the image with its peak at the optical centre as shown in Fig~\ref{fig:gaussian_weight}.
The 2D Gaussian is used to weigh the points closer to the centre of the image higher.
An example of the resulting coloured point cloud is shown in Fig~\ref{fig:coloured_map}.
%

\section{Experimental Method}
\label{sec:Methodology}

\subsection{Detector/Descriptor Pairs}
The keypoints \cite{Lowe2004SIFT, Sipiran2011Harris3D, Zhong2009ISS, Smith1997SUSAN} and features \cite{Frome20043DSC, Rusu2008PFH, Tombari2010USC, Tombari2010Unique, Tombari2011CSHOT} in Table~\ref{tab:detector_descriptor} are selected based on their performance evaluation found in the literature and their accessibility from the open source Point Cloud Library (PCL) \cite{Rusu2011PCL}.
Eight 3D keypoint detectors are evaluated for their robustness and computational complexity.
For each chosen keypoint, 6 descriptors are evaluated in a point cloud alignment pipeline, for a total of 48 unique detector/descriptor pairs.
The parameters for each detector and descriptor are set to default values provided in PCL.
\begin{table}[tb]
    \centering
    \caption{3D detectors / descriptors and their parameters.}
    \label{tab:detector_descriptor}
    \begin{tabular}{c c | c c}
        Detector  & Parameters                     & Descriptor & Size \\
        [1ex]
        \hline
        ISS       & Threshold, Radius Search       & PFH        & 125  \\
        Harris3D  & Threshold, Radius Search       & PFHRGB     & 250  \\
        Lowe      & Threshold, Radius Search       & SHOT       & 352  \\
        Tomasi    & Threshold, Radius Search       & CSHOT      & 1344 \\
        Curvature & Threshold, Radius Search       & 3DSC       & 1980 \\
        Harris6D  & Threshold, Radius Search       & USC        & 1960 \\
        SIFT3D    & Scales, Minimum Contrast       &            &      \\
        SUSAN3D   & Distance and Angular Threshold &            &      \\
        [1ex]
        \hline
    \end{tabular}
\end{table}

\subsection{Keypoint Evaluation}
\label{subsec:Keypoint}
The most important characteristic of a keypoint detector is \textit{repeatability}, defined as the detector's ability to identify the same set of points on different instances of a given model \cite{Salti2011Performance}.
The different instances discussed for keypoint evaluation are viewpoint change and noise corruption.

Optical sensors in underwater applications tend to point downward as they scan structures on the seafloor.
Therefore, keypoint repeatability is evaluated on point clouds rotated about the vertical basis vector of the world frame, $\ura{a_3}$, which is often perpendicular to the seafloor.
A stationary submap, $\mc{S}_{b_{1}}$, resolved in the body frame $\mc{F}_{b_{1}}$ is rotated with angles from $0^{\circ}$ to $180^{\circ}$ with $10^{\circ}$ step to create rotated point clouds.

The robustness of the keypoints is tested on noisy point clouds as well.
Gaussian white noise is injected to $\mc{S}_{b_{1}}$ with $1\sigma$ increments of $0.005$ from $0$ to $0.05$.
The transformed or the noisy submap is defined as the \textit{modified} submap, $\mc{S}_{b_{2}}$.

For each pair of $\mc{S}_{b_{1}}$ and $\mc{S}_{b_{2}}$, two sets of keypoints, ${\mc{K}_{b_{1}} = \{ \mbf{r}^{f_{i}z_{1}}_{b_{1}} \}_{i=1}^{N_{1}}}$ and ${\mc{K}_{b_{2}} = \{ \mbf{r}^{f_{j}z_{2}}_{b_{2}} \}_{j=1}^{N_{2}}}$, are extracted from each submap, respectively.
Similar to \cite{Salti2011Performance}, the keypoints are said to be repeatable if they are approximately coincident,
\begin{align}
    \norm{\mbf{C}_{b_{1}b_{2}} \mbf{r}^{f_{j}z_{2}}_{b_{2}} + \mbf{r}^{z_{2}z_{1}}_{b_{1}} - \mbf{r}^{f_{i}z_{1}}_{b_{1}}} < \epsilon,
\end{align}
where ${\mbf{C}_{b_{1}b_{2}} \in SO(3)}$ and $\mbf{r}^{z_{2}z_{1}}_{b_{1}}$ define the perturbing transformation.
A threshold value ${\epsilon=1\mathrm{e}{-2}}$ is chosen to capture the proximity of corresponding keypoints even in noisy clouds.
Given a set of repeatable keypoints $\mc{RK}$, the relative repeatability $r$ is given by
\begin{align}
    r = \f{\abs{\mc{RK}}}{\abs{\mc{K}_{b_{1}}}}.
\end{align}
The processing time for each keypoint is also evaluated.

\subsection{Feature Evaluation}
\label{subsec:Feature}
\begin{figure}[tb]
    \centering
    \includegraphics[width=\linewidth]{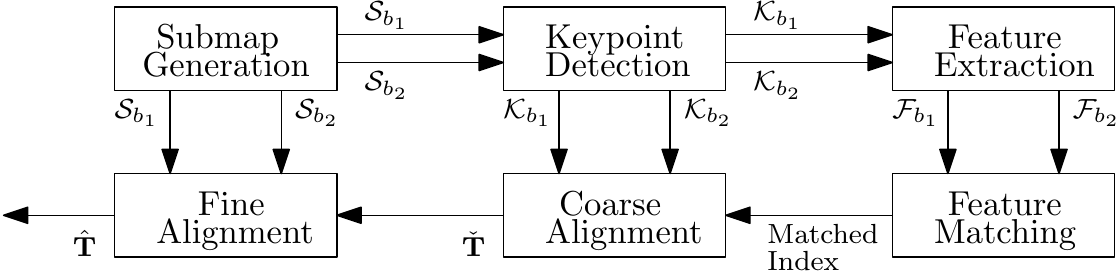}
    \caption{Block diagram of point cloud alignment pipeline.}
    \label{fig:block_diagram}
\end{figure}
Each loop closure produces two submaps at two different vehicle poses from which keypoints are detected and features are extracted.
Extracted features are matched using the $k$-nearest-neighbour algorithm.
The features from source clouds are matched to the target clouds and vice-versa to produce unique matches.

Matched keypoints are used to perform a coarse alignment using \cite{Yang2021TEASER}, which attempts to solve an intractable optimization problem in the presence of a large number of outlier correspondences.
Thus, the coarse alignment algorithm is used as an outlier rejection method.
Feature matching is considered to have failed if the number of inliers is insufficient to allow the coarse alignment algorithm to accurately align the two submaps.

Once the coarse alignment has been performed, a modified recall equation from \cite{Hidalgo2020Evaluation} is used to evaluate the inlier ratio,
\begin{align}
    \mathrm{recall} & = \f{\# \mathrm{inliers}}{\# \mathrm{features}}.
\end{align}
Further, coarse alignment estimates the relative transform between the two submaps ${\mbfcheck{T}^{z_{2}z_{1}}_{b_{2}b_{1}} \in SE(3)}$.
Since ground truth information is not available, the alignment accuracy is evaluated in the matrix Lie algebra $\mathfrak{se}(3)$ as
\begin{align}
    \delta\mbs{\xi}_{i} = \log((\mbfcheck{T}^{z_{2}z_{1}}_{b_{1}b_{2}})_{i}^{-1}(\mbf{T}^{z_{2}z_{1}}_{b_{1}b_{2}})_{i})^{\vee} = \begin{bmatrix}
                                                                                                                                      {\delta\mbs{\xi}^{\phi}_{i}}^{\trans} & {\delta\mbs{\xi}^{\rho}_{i}}^{\trans}
                                                                                                                                  \end{bmatrix}^{\trans},
\end{align}
where $\mbf{T}^{z_{2}z_{1}}_{b_{2}b_{1}}$ is the relative transform obtained from the GPS-aided DVL-INS solution.

\subsection{Full Pipeline Evaluation}
\label{subsec:FullPipeline}
The full pipeline of the algorithm is shown in Fig~\ref{fig:block_diagram}.
The fine alignment is performed based on \cite{Qian2020TWOLATE} and \cite{Hitchcox2020Robust} with the prior pose initialized by the coarse alignment algorithm.
Fine alignment produces a posterior estimate of the relative transform between the two submaps, $\mbfhat{T}^{z_{2}z_{1}}_{b_{1}b_{2}}$.
Since ground truth information is not available, the alignment accuracy is evaluated using the self-consistency error metric of \cite{Roman2007SelfConsistent}.

\section{Results and Discussion}
\label{sec:Results}

\subsection{Keypoint Evaluation}
\label{subsec:KeypointEvaluation}
\begin{figure}[tb]
    \centering
    \subfloat[Rotation\label{kp_result_a}]{%
        \includegraphics[width=0.9\linewidth,clip=true,trim={0.75cm 0.25cm 0.75cm 2cm}]{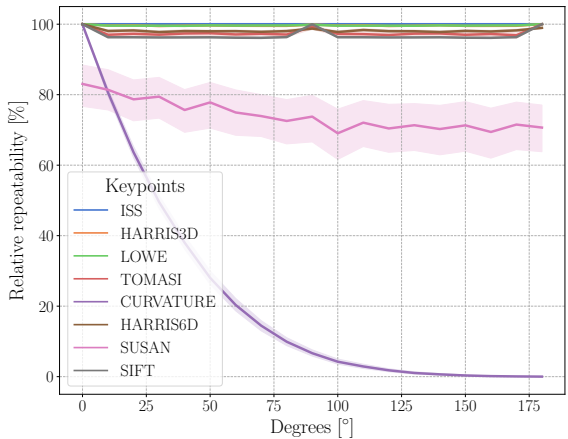}} \\
    \subfloat[Point cloud noise\label{kp_result_b}]{%
        \includegraphics[width=0.9\linewidth,clip=true,trim={0.75cm 0.25cm 0.75cm 2cm}]{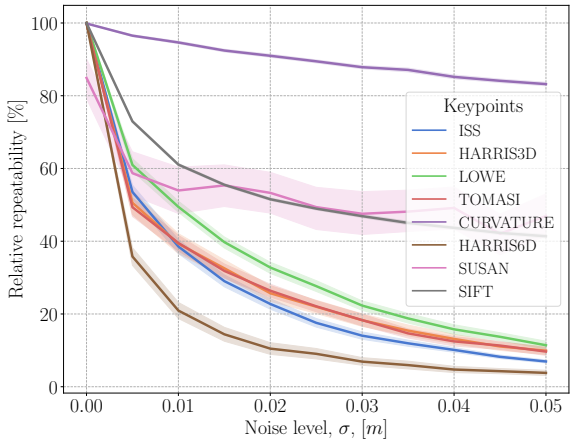}}
    \caption{Keypoint repeatability results}
    \label{fig:kp_result}
\end{figure}

The first set of experiments investigates which keypoints are most robust to rotation and point cloud noise.
The relative repeatability of each keypoint in all underwater scenes is shown in Fig~\ref{fig:kp_result}.
Solid lines represent the mean value, and the banded areas represent a $95\%$ confidence level.
ISS keypoints have the highest mean repeatability with the lowest variance for rotated point clouds, followed by Harris3D and Lowe.
ISS, Harris3D, Lowe, Tomasi, Harris6D and SIFT are rotation invariant.
Curvature and SUSAN vary with rotation as the repeatability decreases with the amount of rotation.
However, the Curvature method shows the highest robustness to the point cloud noise followed by SUSAN and SIFT.
Harris6D is the most susceptible to point cloud noise.
Further, Table~\ref{tab:percent_success1} shows that ISS and SIFT are the most versatile keypoints that can perform with various descriptors.

The average time taken to detect the keypoints is shown in Table~\ref{tab:kp_time_analysis}.
SIFT takes the longest to detect at around $\SI{1.2}{\second}$ per submap, followed by Harris6D and ISS.
SUSAN and Curvature take the least time to detect at around $\SI{0.002}{\second}$ and $\SI{0.004}{\second}$, respectively.
As SUSAN yields the lowest detection time with relatively high robustness to point cloud noise and rotations, it may be suitable for real-time applications.
\begin{table}[tb]
    \centering
    \caption{Keypoint detection time analysis in milliseconds. The abbreviated column names are H3D: Harris3D, CU: Curvature, and H6D: Harris6D.}
    \label{tab:kp_time_analysis}
    \begin{tabular}{c c c c c c c c}
        ISS  & H3D  & Lowe & Tomasi & CU   & H6D  & SIFT & SUSAN         \\
        [1ex]
        \hline
        16.3 & 13.9 & 13.7 & 12.2   & 4.02 & 24.1 & 1213 & \textbf{2.51} \\
        \hline
    \end{tabular}
\end{table}
\subsection{Feature Evaluation}
\label{subsec:FeatureEvaluation}
\begin{figure*}[tb]
    \centering
    \subfloat[Rotation errors\label{feat_result_a}]{%
        \includegraphics[width=0.33\textwidth,clip=true,trim={0.5cm 1cm 2cm 1cm}]{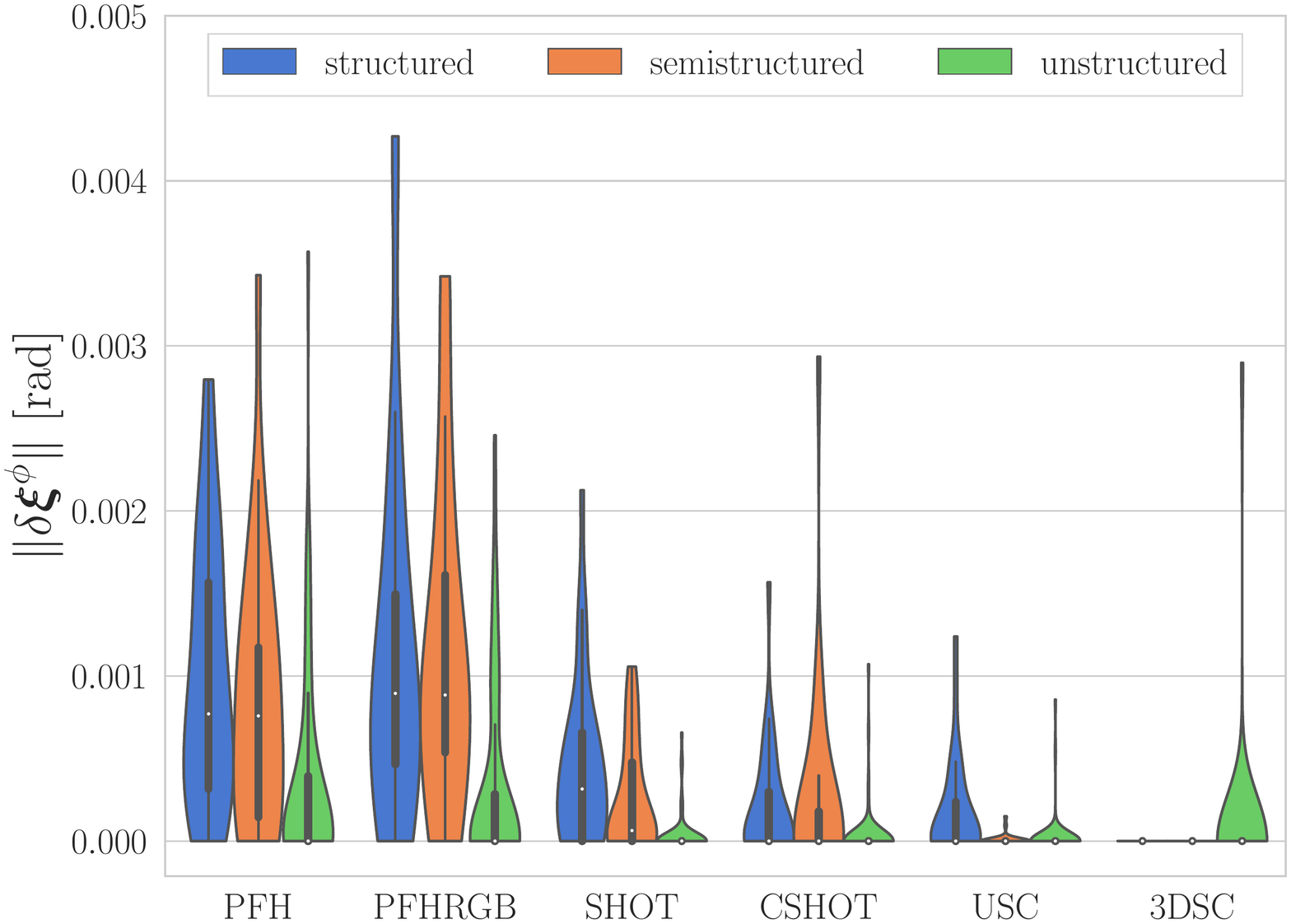}}
    \hfill
    \subfloat[Translation errors\label{feat_result_b}]{%
        \includegraphics[width=0.33\textwidth,clip=true,trim={0.5cm 1cm 2cm 1cm}]{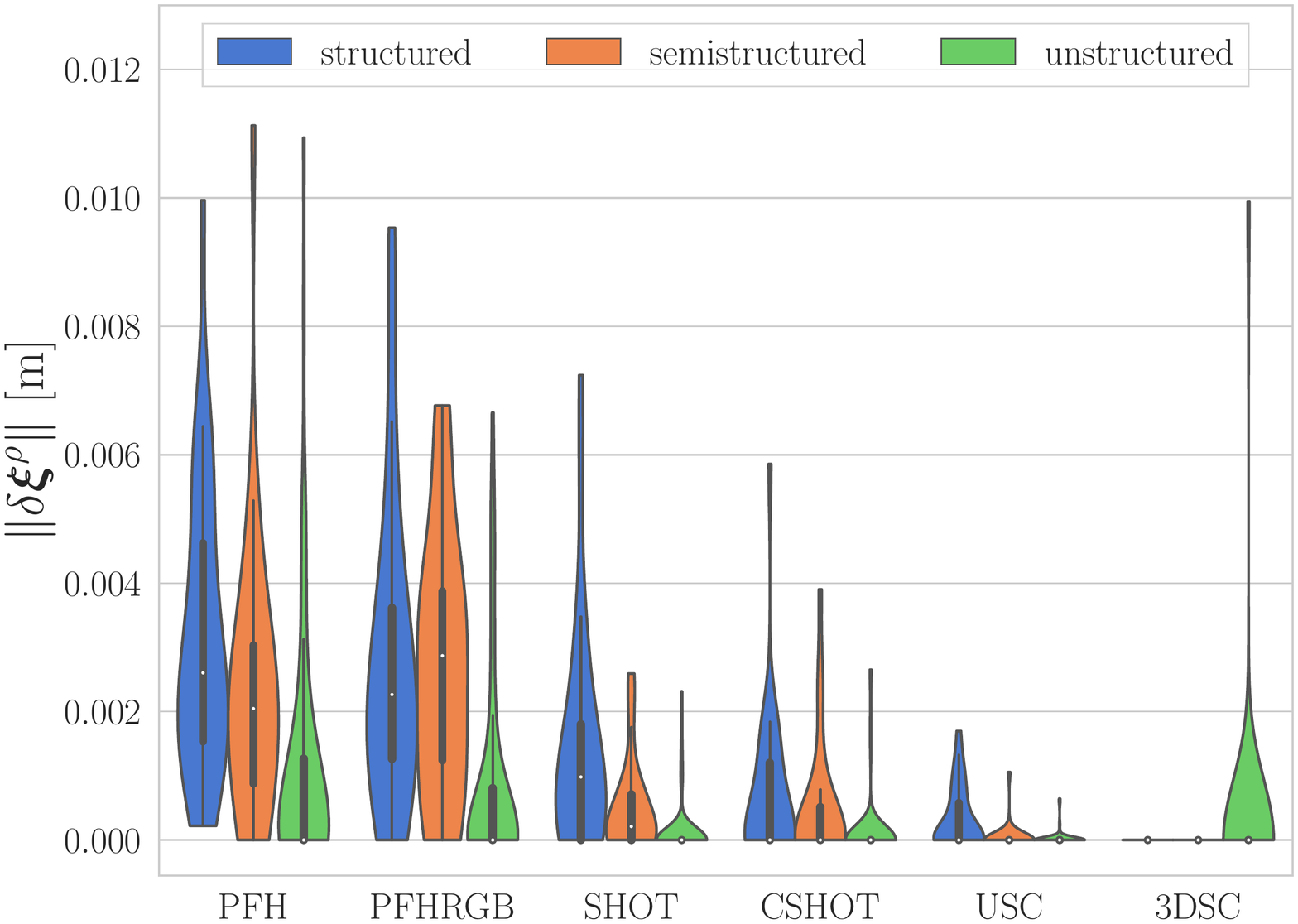}}
    \hfill
    \subfloat[Recall\label{feat_result_c}]{%
        \includegraphics[width=0.33\textwidth,clip=true,trim={0.5cm 1cm 2cm 1cm}]{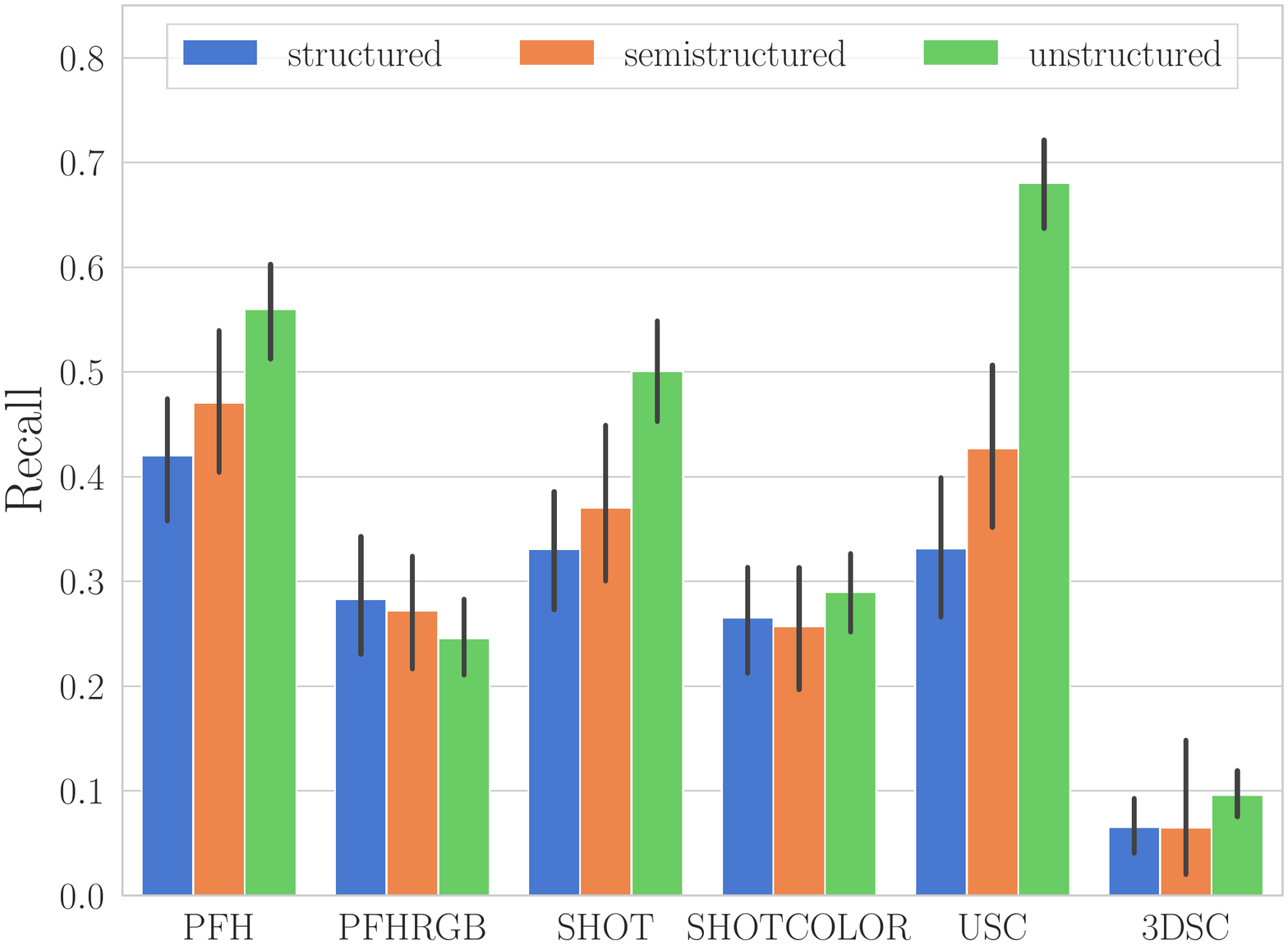}}
    \caption{Coarse alignment errors for ISS keypoints, grouped by category of underwater environment.}
    \label{fig:feat_result}
\end{figure*}

\begin{table}[tb]
    \caption{Percent success in feature matching for every descriptor pair}
    \label{tab:percent_success1}
    \begin{tabular}{c | c c c c c c}
               & PFH          & PFHRGB        & SHOT         & CSHOT & 3DSC & USC \\
        [1ex]
        \hline
        ISS    & \textbf{100} & \textbf{95.8} & \textbf{100}
               & \textbf{100} & 40.7          & 99.2                              \\
        H3D    & 94.9         & 83.9          & 84.7
               & 75.4         & 21.2          & 98.3                              \\
        Lowe   & 99.2         & 91.5          & 85.6
               & 82.2         & 23.7          & 99.2                              \\
        Curv   & 64.4         & 48.3          & 36.4
               & 30.5         & 19.5          & 69.5                              \\
        Tomasi & 90.7         & 75.4          & 78.0
               & 69.5         & 21.2          & 95.6                              \\
        H6D    & 96.6         & 88.1          & 88.1
               & 80.5         & 23.7          & 99.2                              \\
        SUSAN  & 41.5         & 44.1          & 24.6
               & 30.5         & 17.8          & 39.8                              \\
        SIFT   & 96.6         & 89.8          & 83.9
               & 82.2         & \textbf{53.4} & \textbf{100}                      \\
        [1ex]
        \hline
    \end{tabular}
\end{table}
\begin{table}
    \caption{Percent success in feature matching for different scenes}
    \label{tab:percent_success2}
    \begin{tabular}{c | c c c c}
        Feature & Structured     & Semi-structured & Unstructured   & Total          \\
        [1ex]
        \hline
        PFH     & 93.30          & \textbf{92.71}  & 79.55          & 85.49          \\
        PFHRGB  & 88.84          & 86.46           & 68.75          & 77.12          \\
        SHOT    & 54.91          & 72.92           & 80.11          & 72.67          \\
        CSHOT   & 50.45          & 70.83           & 75.95          & 68.86          \\
        3DSC    & 51.34          & 17.19           & 21.40          & 27.65          \\
        USC     & \textbf{95.09} & 91.14           & \textbf{83.14} & \textbf{87.61} \\
        [1ex]
        \hline
    \end{tabular}
\end{table}

The features are evaluated with every detector/descriptor combination in all 118 loop-closure events classified into 3 different underwater scenes.
For structured scenes, USC has the highest success rate in feature matching followed by PFH and PFHRGB as shown in Table~\ref{tab:percent_success2}.
USC also has the highest success rate for the unstructured environments, while all other descriptors except 3DSC show relatively similar performance.
Overall, USC is the most robust keypoint in different underwater scenes with the highest success rate.

The alignment errors are plotted and compared in Fig~\ref{fig:feat_result}.
The rotation and translation errors exhibit similar patterns for each descriptor.
3DSC seems to perform the best with coarse alignment, but it has the lowest recall value shown in Fig~\ref{feat_result_c}, indicating that the number of inliers is too low to accurately perform coarse alignment.
Thus, USC performs the best in terms of the alignment accuracy, followed by CSHOT.

Feature matching has a greater tendency to fail in unstructured environments, as shown in Table~\ref{tab:percent_success2}, yet successful alignments in unstructured environments produce the lowest alignment error.
This means that once the correspondences are found, the coarse alignment succeeds with very low transformation error.
Features tend to have higher chances of finding matches in structured scenes, but with higher alignment error.
This may be because the coarse alignment algorithm is unable to find a sufficient number of inliers.

Despite USC delivering the highest performance in terms of the robustness to environments and the alignment accuracy, USC is the second-slowest in processing time.
Thus, USC may not be the most adequate descriptor for real-time applications.
As shown in Table~\ref{tab:feat_time_analysis}, SHOT and CSHOT outperform in overall processing by a significant margin, and thus may be considered as candidates for real-time applications.
PFH and PFHRGB can also be considered as candidates for increased robustness at the expense of longer processing times.
\begin{table}[tb]
    \centering
    \caption{Feature processing time analysis in milliseconds. The abbreviated column names are CA: coarse alignment, and FA: fine alignment.}
    \label{tab:feat_time_analysis}
    \begin{tabular}{c c c c c c}
        Feature & Extraction     & Matching       & CA             & FA             & Total          \\
        [1ex]
        \hline
        PFH     & 124.7          & \textbf{5.619} & 50.22          & 266.0          & 446.5          \\
        PFHRGB  & 205.7          & 7.638          & \textbf{48.64} & 269.1          & 531.1          \\
        SHOT    & \textbf{14.97} & 9.686          & 56.92          & 219.3          & \textbf{300.9} \\
        CSHOT   & 17.04          & 23.38          & 58.11          & \textbf{203.7} & 302.2          \\
        3DSC    & 712.1          & 32.68          & 53.86          & 224.5          & 1023           \\
        USC     & 513.7          & 33.06          & 51.04          & 227.2          & 825.0          \\
        \hline
    \end{tabular}
\end{table}
Unfortunately, the experimental results produced no evidence that adding texture information to the point cloud scan improves alignment robustness.
This may be due to the greyish, monochromatic colours in the images caused by low illumination underwater.
This can possibly be improved with the development of a better colour levelling algorithm.

\subsection{Full Pipeline Evaluation}
\begin{figure}[tb!]
    \centering
    \subfloat[Initial alignment errors\label{wolate_result_a}]{%
        \includegraphics[width=\linewidth,clip=true,trim={0.5cm 0.25cm 0cm 0.5cm}]{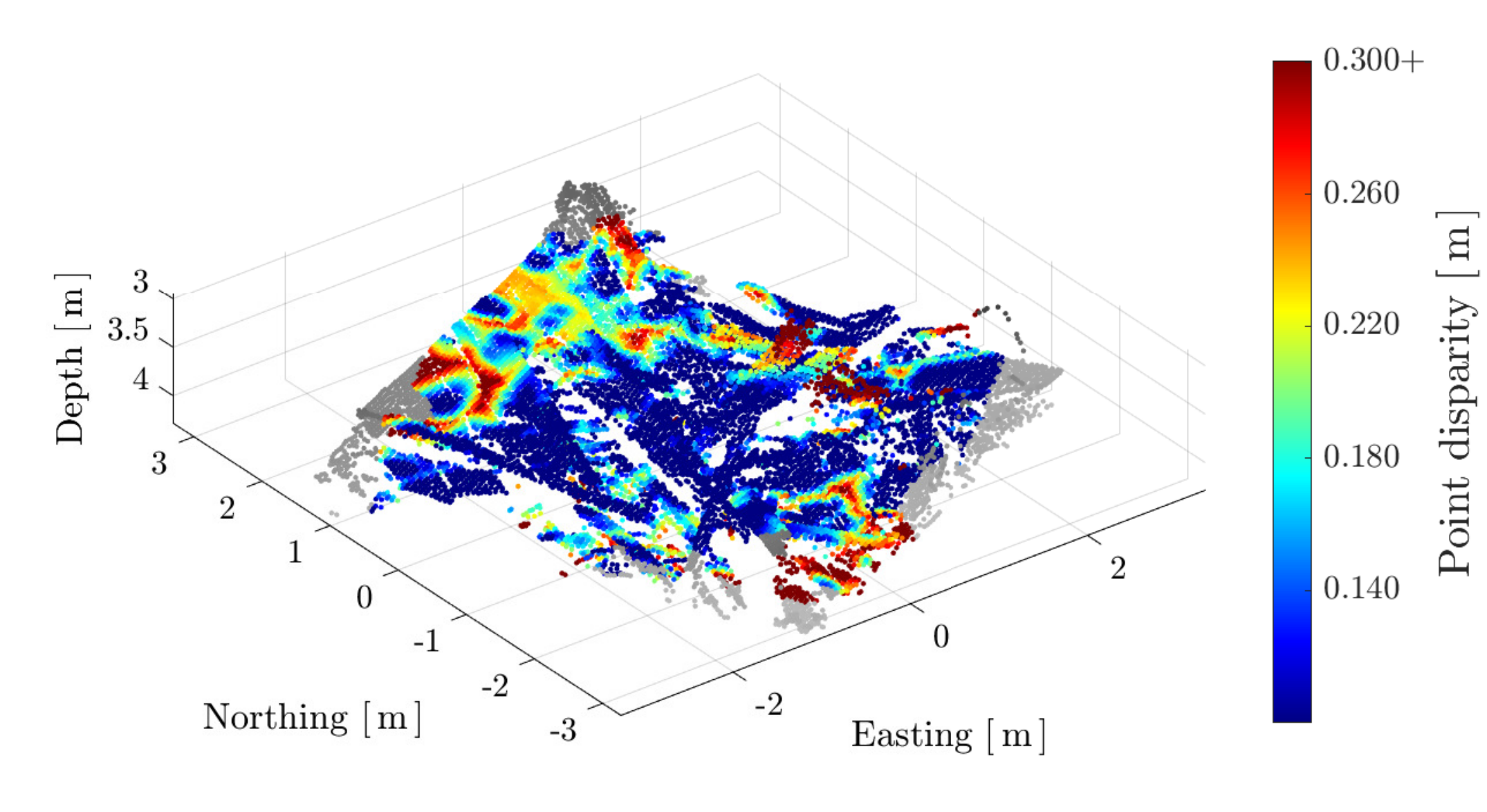}}
    \hfill
    \subfloat[Errors after coarse alignment\label{wolate_result_b}]{%
        \includegraphics[width=\linewidth,clip=true,trim={0.5cm 0.25cm 0cm 0.5cm}]{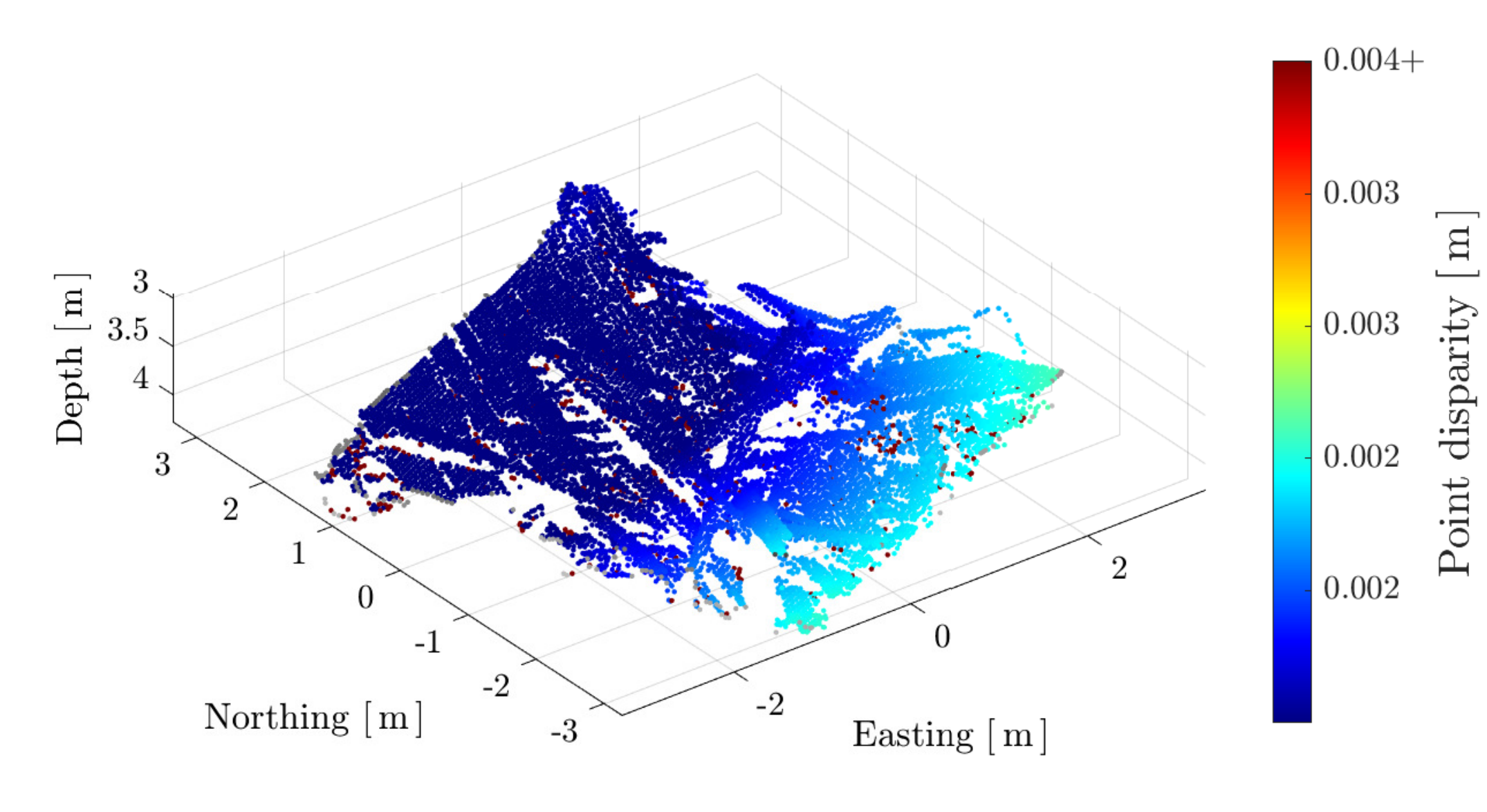}}
    \hfill
    \subfloat[Errors after fine alignment\label{wolate_result_c}]{%
        \includegraphics[width=\linewidth,clip=true,trim={0.5cm 0.25cm 0cm 0.5cm}]{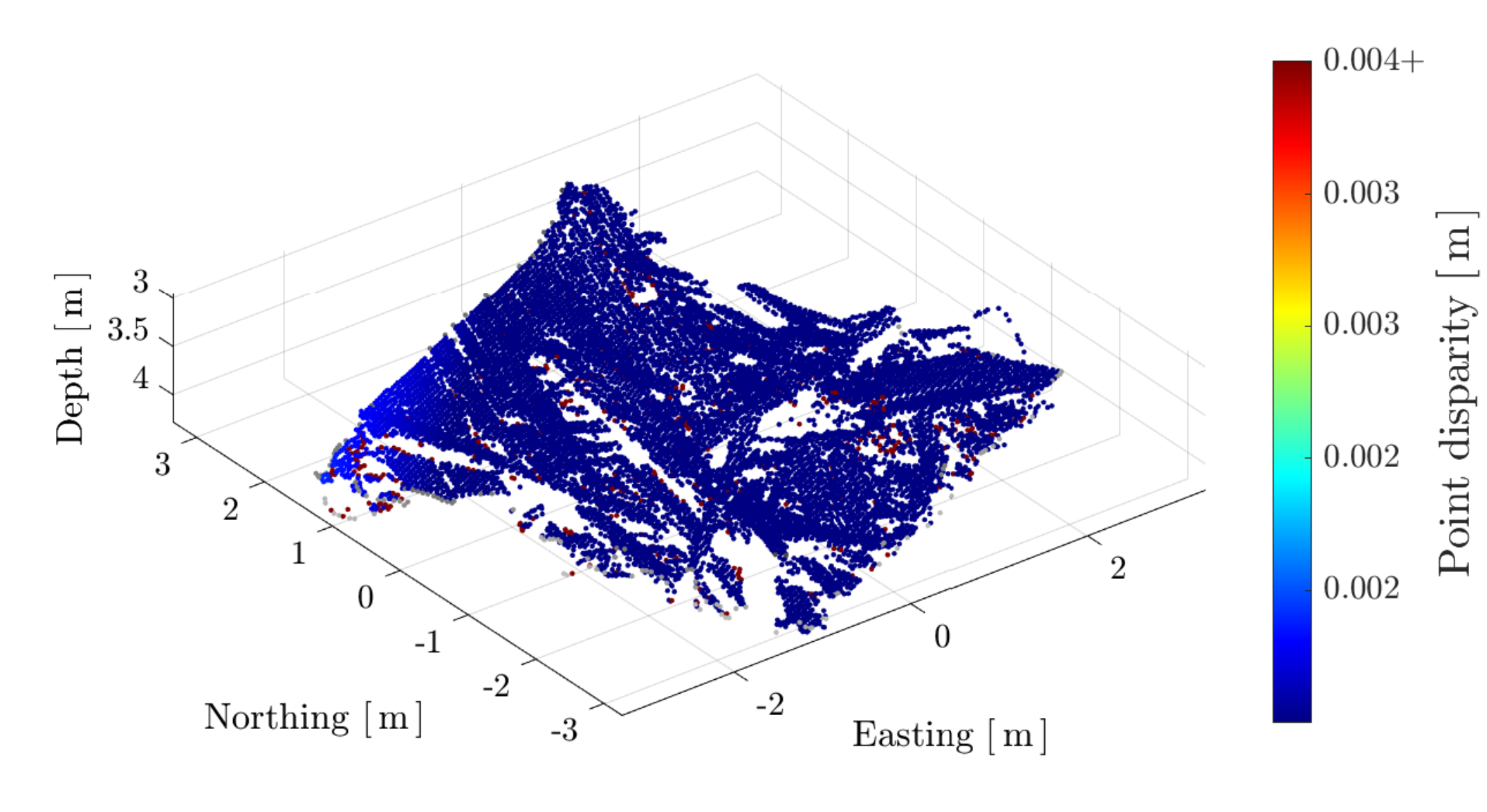}}
    \hfill
    \caption{Consistency-based errors \cite{Roman2007SelfConsistent} are computed on a loop closure from Shipwreck B data with ISS keypoints and USC descriptors.}
    \label{fig:wolate_result}
\end{figure}

The full point cloud alignment pipeline illustrated in Fig~\ref{fig:wolate_result} is run with a combination of ISS keypoints and USC descriptors, as this combination exhibits the highest robustness to environments changes and the highest accuracy in the coarse alignment algorithm.
Fig~\ref{fig:wolate_result} shows the initial alignment errors, the errors after coarse alignment, and the errors after fine alignment.
The errors after coarse alignment are low enough to properly initialize the ICP algorithm \cite{Qian2020TWOLATE}, and align the two submaps together.
%

\section{Conclusion}
\label{sec:Conclusion}


This paper evaluates the effectiveness of different detector/descriptor pairs for point cloud data collected in novel and challenging underwater environments.
Additionally, a novel method has been proposed for colourizing subsea point cloud scans with images collected from an underwater colour camera.
This allows descriptors incorporating colour information to be included as part of the study.

The study shows that ISS, Lowe, and Harris3D outperform other keypoints in terms of point cloud noise and rotation, while SUSAN outperforms others in terms of processing time.
The USC feature outperforms other features in matching rate and alignment accuracy followed by PFH and PFHRGB.
Interestingly, features with colour information do not outperform their geometric counterparts, possibly due to low illumination and contrast in the underwater environment.
The evaluation of newly developed machine learning-based detectors and descriptors is an interesting avenue for future research.

    {\AtNextBibliography{\small}
        \printbibliography}
\end{document}